# Fractional Vegetation Cover Estimation using Hough Lines and Linear Iterative Clustering

Venkat Margapuri
Department of Computer Science
Kansas State University
Manhattan, KS
marven@ksu.edu

Trevor Rife
Department of Plant Pathology
Kansas State University
Manhattan, KS
trife@ksu.edu

Chaney Courtney
Department of Plant Pathology
Kansas State University
Manhattan, KS
chaneylc@ksu.edu

Brandon Schlautman
The Land Institute
Salina, KS
schlautman@landinstitute.org

Kai Zhao
Department of Computer Science
Kansas State University
Manhattan, KS
kaizhao@ksu.edu

Mitchell Neilsen
Department of Computer Science
Kansas State University
Manhattan, KS
neilsen@ksu.edu

*Abstract*— A common requirement of plant breeding programs across the country is companion planting -- growing different species of plants in close proximity so they can mutually benefit each other. However, the determination of companion plants requires meticulous monitoring of plant growth. The technique of ocular monitoring is often laborious and error prone. The availability of image processing techniques can be used to address the challenge of plant growth monitoring and provide robust solutions that assist plant scientists to identify companion plants. This paper presents a new image processing algorithm to determine the amount of vegetation cover present in a given area, called fractional vegetation cover. The proposed technique draws inspiration from the trusted Daubenmire method for vegetation cover estimation and expands upon it. Briefly, the idea is to estimate vegetation cover from images containing multiple rows of plant species growing in close proximity separated by a multi-segment PVC frame of known size. The proposed algorithm applies a Hough Transform and Simple Linear Iterative Clustering (SLIC) to estimate the amount of vegetation cover within each segment of the PVC frame. The analysis when repeated over images captured at regular intervals of time provides crucial insights into plant growth. As a means of comparison, the proposed algorithm is compared with SamplePoint and Canopeo, two trusted applications used for vegetation cover estimation. The comparison shows a 99% similarity with both SamplePoint and Canopeo demonstrating the accuracy and feasibility of the algorithm for fractional vegetation cover estimation.

## I. Introduction

Companion planting is the idea of growing multiple species of plants in close proximity so that they reap mutual benefits, such as improved crop yield, soil quality and pest control. One of the foremost instances of companion planting in North America is that of 'three sisters' pioneered by the Native Americans. The plant varieties of beans, corn and squash constitute the 'three sisters' where their cultivation in close proximity led to the benefits of shelter and growth support for plants in addition to improved soil quality and decreased soil erosion [11]. Other combinations of plant varieties that complement each other's growth include artichoke and cucumber, beetroot and onion, and tomatoes and carrot whereas the combinations of parsnip and carrot, potatoes and pumpkin, and peas and garlic are instances of poor companions that deter one another's growth and development [19]. Hence, the discovery of companion plant species that are able to co-exist is of significant value to the agricultural sector. A key metric that helps estimate the growth of plant species is Fractional Vegetation Cover (FVC). FVC is defined as the percentage of the ground surface covered by vegetation elements from the overhead perspective [2]. Different techniques such as Point-and-Line cover Estimation, Plot-based cover estimation, Ocular-based estimation and semi-quantitative ocular based estimation are available to estimate FVC at a certain location. However, the techniques currently available are labor-intensive, stochastic and subjective in nature. In addition, the techniques are prone to overestimating the vegetation cover due to field survey design [16]. The need for a technique that precisely analyzes the amount of FVC persists. The article proposes a technique to precisely estimate the amount of FVC in an area from images by drawing inferences from and extending upon the Daubenmire method, a semi-quantitative ocular-based FVC estimation technique. While current techniques estimate the percentage of FVC within images, the proposed technique provides an estimate on the area occupied by FVC in metric units in addition to percentage. The precise estimation of the FVC helps the plant scientists make discoveries of plant varieties that make good companions and offer well-informed suggestions to the agrarian community.

## II. Related Work

The development of vegetation cover rating scales started in the early 1900s. The most widely accepted scale at the time was the rating scale proposed by Braun-Blanchet [7]. The scale involved estimating and classifying the vegetation cover within a certain area into one of five cover classes from one to five where one indicated cover less than five percent, two indicated cover between five and twenty-five percent, three meant cover between 25% and 50%, four meant cover between 50% and 75%, and five meant cover between 75% and 100%.

RF Daubenmire [7] proposed the canopy coverage method in 1959. It was a semi-quantitative ocular based estimation technique and regarded as one of the most accurate methods for vegetation cover analyses over the years. It involved meticulously placing a 20- x 50- cm quadrat along a tape on permanently located transects and classifying the amount of vegetation cover into one of six cover classes. The cover classes are similar to Braun-Blanchet's classes with the difference being that cover class five was divided into two classes where cover class five indicated vegetation cover between 75% and 95% and class six indicated vegetation cover between 95% and 100%.

Booth, Cox and Berryman [8] proposed SamplePoint, a free vegetation cover estimation tool from digital images using manual point sampling. The application shows users multiple single-pixel sample points (defaults to 100) on the image and allows them to classify the pixel as belonging to one of nine categories. The application uses the classifications made by the users to identify the percentage of pixels belonging to each of the nine categories. The results are output to an excel spreadsheet and have been comparable to the results from the most accurate field experiments for vegetation cover estimation.

Systat Software Inc. [27] provides specialized scientific software products for research in fields such as environmental sciences, life sciences and engineering. SigmaScan Pro 5.0, an image processing software tool developed by Systant Software Inc. may be used to estimate the percentage of pixels that belong to vegetation from digital imagery. While the tool was not developed for vegetation cover estimation, it is able to be adapted for the use case.

Patrignani and Ochsner [9, 24] developed a software tool named Canopeo to estimate the fractional green canopy cover from digital images. The tool was developed using Matlab and red-to-green (R/G), blue-to-green (B/G) and excess green index (2G-R-B). Desktop and mobile versions of the tool for android and iOS are available as free downloads. The tool provides a percentage estimate of the amount of green cover within an image and provides a grayscale image highlighting the green cover. However, the tool doesn't provide the vegetation cover estimate in metric units making it hard to reproduce results among experiments across individuals due to difference of height of image capture.

Louhaichi. M., Hassan, S., Clifton, K., & Johnson, D. E. [20, 21] proposed VegMeasure, a software tool that processes digital imagery collected in a specialized manner known as Digital Vegetative Charting Technique (DVCT). VegMeasure provides classification of imagery and measures change over time. However, DVCT requires a digital camera with built-in GPS that can be mounted to a stand so images are captured from a fixed height with lens pointed orthogonal to the surface. VegMeasure requires that the camera, its height and orientation be kept constant throughout the image capture process for its estimation and analysis.

Laliberte et. al [18] demonstrated that object-based image analysis upon the conversion of RGB scale images to IHS (intensity-hue-saturation) scale images is a viable approach for estimating total cover of vegetation, bare soil, and fractional components of green and senescent vegetation. The image analysis tool used for the experiments was eCognition where the image was segmented into homogenous areas based on three parameters: scale, color and shape.

## III. ALGORITHM

The breeders at the Land Institute in Salina, KS aim to estimate FVC across multiple plant varieties with the goal to identify companion plant species and make quality recommendations may be made to the farming community. The design of the experiment is inspired by the Daubenmire canopy coverage method and visually depicted by Figure 1. Briefly, the Daubenmire technique involves using a 20 cm x 50 cm quadrat around vegetation cover along a tape on permanently located transects. The vegetation within the quadrat is measured by trained plant scientists based on ocular estimation. Since estimating the precise amount of plant cover from ocular observation is farfetched, plant scientists classify (estimate) the plant cover into one of six cover classes labelled one through six, as defined by Daubenmire. Class one indicates crop cover less than five percent, class two indicates crop cover between five and twenty-five percent, class three indicates crop cover between 25% and 50%, class four indicates crop cover between 50% and 75%, class five indicates crop cover between 75% and 95%, and class six indicates crop cover between 95% and 100%. As part of the experiment, a PVC frame of known dimensionality is dropped across multiple rows of crops as shown in Figure 2(a) and the amount of vegetation cover within each of the segments of the PVC frame is estimated independently at regular intervals over a stipulated time frame. Since the experiment progresses over time, accurate readings at each time step are essential to estimate the growth of vegetation. An algorithm that leverages image processing and analysis is proposed to eliminate the need for ocular estimation of FVC and measure FVC from digital images with precision. The development of the algorithm is

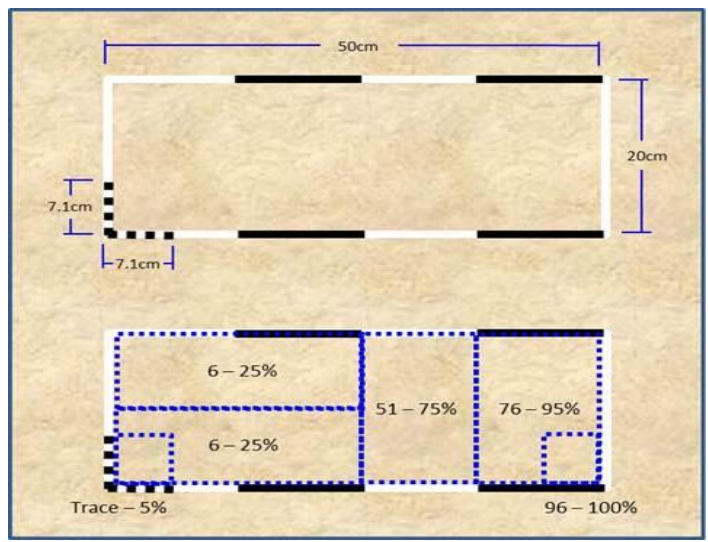

Figure 2: Daubenmire Quadrat with Ground Cover Classes [7]

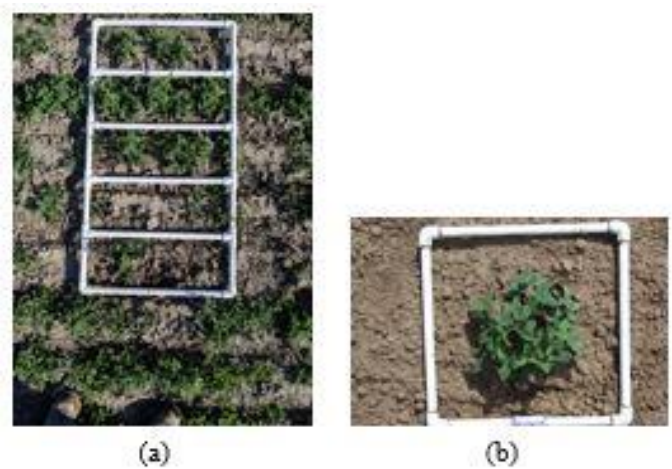

Figure 1: (a) Vegetation Cover within Five-Segment PVC Frame (b) Kura Plant within One-Segment PVC Frame

performed on images from two datasets where one of the datasets comprises 33 images that capture alfalfa at a resolution of 3024 x 4032 pixels and the other comprises 177 images that capture a one-segment PVC frame laid around plants of Kura at a resolution of 5184 x 3456 pixels, as shown in Figure 2. Broadly, the algorithm is a three-step process where in the first and second steps are the detection and extraction of the PVC frame and vegetation cover from the image respectively. The final step is the estimation of the amount of vegetation cover within each of the segments of the PVC frame. The algorithm is developed in Python using the OpenCV [10] image processing library. The code for the algorithm is available at: https://github.com/marven22/Fractal-Vegetation-Cover-Estimation.git

*A. Noise Removal and PVC Frame Extraction from Image*

In addition to vegetation cover within the PVC frame, the image consists of area outside the PVC frame. The area outside the PVC frame is irrelevant in terms of vegetation cover estimation within the grid. Hence, the first step of the process is the meticulous determination of the region of interest i.e. the PVC frame and pixels within the PVC frame. The image processing concepts that are relevant to the process are HSV color space, Perspective Transform and 'bitwise and' operation.

**HSV Color Space:** HSV refers to Hue, Saturation and Value which make up the co-ordinates for the color space, as shown in Figure 3. It is a cylindrical color space where the radius represents Saturation, the vertical axis represents Value and the angle represents Hue. Intuitively, Hue is the dominant color visible to an observer, Saturation is the amount of white light mixed with a hue and Value is the chromic notion of intensity. As Value decreases, the color gets closer to black whereas the intensity of the color increases as Value increases.

**Perspective Transform:** Perspective Transform corrects the perspective of an image to bird's eye or top view. In other words, it helps view the image as if it were captured with the camera held orthogonal to the surface. The application of the transform immensely helps reduce skew in images captured using freehand. The downside to images being skewed is that objects in skewed images appear to be of a different size and orientation in comparison to their true dimension. In the context of the current algorithm, the application of the perspective transform to the images helps reduce skew, thereby making the process of identifying the PVC frame easier.

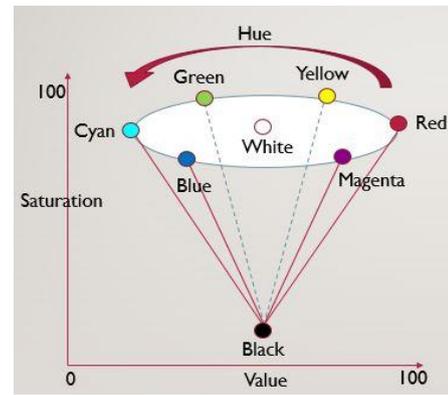

Figure 3: HSV Color Space

**Bitwise AND Operation:** The 'bitwise and' operation is used to extract a specific region from an image using a mask. The mask is a grayscale image provided to indicate the region of the image required to be extracted. Typically, the pixels of the mask are made to be either white (255) or black (0). The original image is also converted to grayscale and all the regions of the image that are not a black (0) pixel are set to white (255). The 'bitwise and' operation operates on a pixel-by-pixel basis where it identifies the common regions between the two and sets them to white (255).

The algorithmic steps to remove noise and extract the PVC frame are described as follows and the results produced after each step are shown in Figure 4.

1. Convert the original image in RGB color space to HSV color space and identify the pixels that belong to the PVC

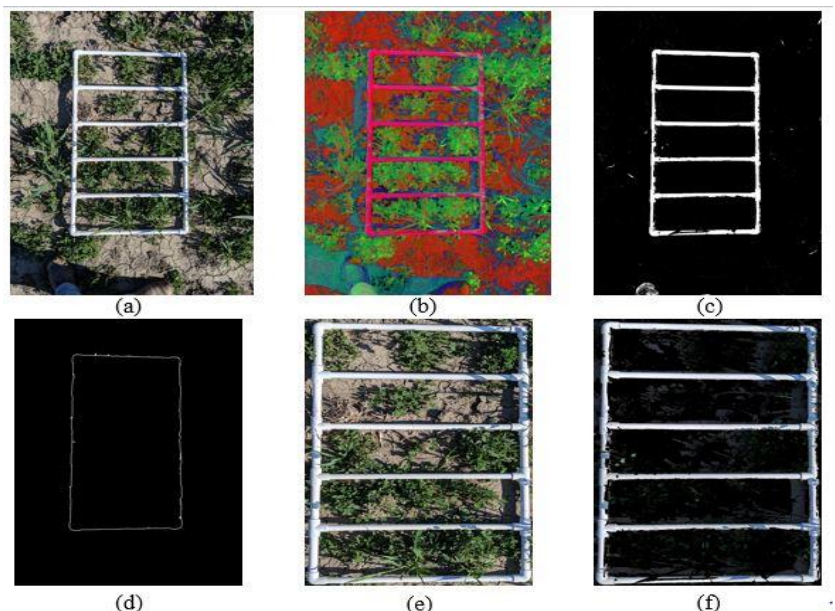

Figure 4: (a) Original Image (b) Original Image in HSV Color Scheme (c) Grayscale Image of PVC Frame

(d) Contour of the PVC Frame (e) Original Image without Noise (f) Extracted PVC Frame from Original Image

frame using the lower and upper ranges of (70, 0, 110) and (180, 255, 255) for (hue, saturation, value) respectively.

2. Convert the detected pixels of the PVC frame to grayscale and apply a median blur on the image to smoothen the image, if necessary.

3. Perform canny edge detection on the image and detect contours on the edges obtained.

4. Identify the largest contour of the contours detected. The idea is that the largest contour runs along the circumference of the PVC frame since the contours of each of the individual segments are significantly smaller in comparison to the outer contour which encapsulates the PVC frame.

5. Plot the minimum area rotated rectangle around the contour of the PVC frame.

6. Perform a 'bitwise and' operation on the input image using the minimum area rotated rectangle as the mask. Doing so precisely identifies the location of the PVC frame on the image. The region of the image within the PVC frame is the only portion of the image to be processed for vegetation cover estimation.

7. Apply perspective transform on the image to view the image from the top view. It corrects skew and makes the task of detecting segments on the PVC grid easier.

8. Convert the output of step 7 to HSV format and apply the lower and upper ranges mentioned in step 1 to precisely extract the PVC frame.

### B. Vegetation Cover Detection using Simple Linear Iterative Clustering (SLIC)

On the region of interest identified in the previous step, SLIC is applied to segment the image into multiple super-pixels. To elaborate, the process of partitioning an image into multiple segments by assigning a label to every pixel in the image is called image segmentation. The idea is that segments with the same label are perceptually similar and share common characteristics. Image segmentation helps to identify similar regions in images and extract semantic information from them. While segmenting images by pixel is an intuitive approach, segmenting by super-pixel is more optimal. A super-pixel is defined as a group of pixels that perceptually belong together by sharing common characteristics such as pixel color, intensity and other low-level properties. Modern cameras result in images of high resolution. While the level of detail within the images is excellent, it overburdens the image processing algorithms used to process the images resulting in longer runtimes. The key benefit to using super-pixels over pixels is that they shrink the pixel space required to be processed by the algorithm and in turn, lead to shorter runtimes without compromising accuracy. In addition, a single pixel by itself does not provide any semantic information whereas a super-pixel provides semantic information that helps make key discoveries within images.

**Simple Linear Iterative Clustering (SLIC):** SLIC is an image processing technique that clusters pixels to generate super-pixels based on their color similarity and proximity in the image plane. SLIC uses the five dimensional [LABXY] space for the clustering where [LAB] is the pixel color vector in CIELAB color space and [XY] is the pixel position on the XY plane [14]. The CIELAB color space, as shown in Figure 5, is defined by the International Commission on Illumination (CIE) to improve upon the fact that the conventional RGB color space only allows for the representation of 40% of the colors that the human eye can perceive. The L channel represents lightness which ranges from black to white, the A channel represents the value on the axis that ranges from green to red and the B channel represents the value on the axis that ranges from blue to yellow. A key characteristic of the CIELAB color space is that its dimensions are non-linearly scaled in the sense that spatial distance between two colors corresponds to perceptual distance and is uniform across the color space. For instance, consider two pairs of different colors that appear to be equally similar such as blue and light blue, red and light red. When Euclidean distance is computed between the two pairs of colors in the CIELAB color space, it is observed the Euclidean distance between both pairs of colors is the same.

In order to use the Euclidean distance measure in the five dimensional [LABXY] space, the spatial distance between two points is required to be normalized since it is subject to image size. Consider an image containing N pixels and let K be the number of super-pixels in the segmented image. The size of each super-pixel is N/K and a super-pixel center is present at every grid interval $S = \sqrt{N/K}$. The center of each of the super-pixels $C_k = [L_k, A_k, B_k, X_k, Y_k]$ where $k = [1, K]$ is chosen at regular grid intervals S. The spatial extent of each of the super-pixels is approximately $S^2$ i.e. the area of a super-pixel and the pixels that are associated with any super-pixel center lie within a 2S x 2S area around the super-pixel center on the XY plane [3]. Euclidean distances in CIELAB color space are perceptually meaningful for small distances but not large distances. The distance measure $D_s$ used in SLIC is the aggregate of the LAB distance and XY plane distance normalized by the grid interval S. The mathematical notations are as follows:

LAB distance, $D_{lab} = \sqrt{(l_k - l_i)^2 + (a_k - a_i)^2 + (b_k - b_i)^2}$

XY plane distance, $D_{xy} = \sqrt{(x_k - x_i)^2 + (y_k - y_i)^2}$

SLIC distance measure, $D_s = D_{lab} + (m/S) * D_{xy}$ where m is the factor that controls compactness of the super-pixel. The greater the value of m, the more compact the cluster.

The SLIC algorithm is similar to any other clustering algorithm in the sense it begins by sampling K regularly spaced cluster centers and moving them to seed location corresponding to the lowest gradient position in a 3 x 3 neighborhood. Image gradients are computed as $G(x, y) = \|I(x + 1, y) - I(x - 1, y)\|^2 + \|I(x, y + 1) - I(x, y - 1)\|^2$ where I

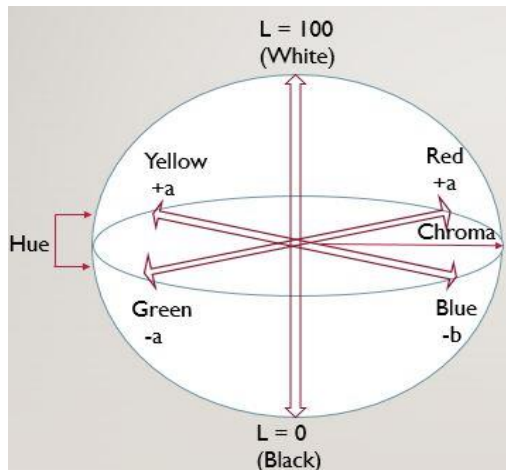

Figure 5: CIELAB Color Space

(x, y) is the LAB vector corresponding to the pixel at position (x, y) and ||.|| is the $L_2$ norm [3]. Each pixel in the image is associated with the nearest cluster center whose search area overlaps the pixel. After all the pixels are associated with the nearest cluster center, a new center is computed as the average LABXY vector of all the pixels belonging to the cluster. The process is iteratively repeated until convergence.

The regions that indicate plant cover are identified by inferring from the semantics of the super-pixels put out by SLIC. The steps in the process are described as follows.

1. Segment the image into a pre-defined number of super-pixels using the SLIC function from the OpenCV image processing library. It is observed that segmenting the image into 300 super-pixels identifies the plant cover best although the algorithm may execute faster at the expense of accuracy if the image is segmented into fewer super-pixels. The image segmented into super-pixels is shown in Figure 6(a).

2. Find the average color value within each of the super-pixels and assign it as the label that represents the super-pixel. It ensures that a given super-pixel has only one color.

3. Find unique labels within the image and apply a filter that identifies the plant cover. An observation that is made based on experimentation is that the super-pixels that consist of plants are green. Hence, all the green labels are identified and a grayscale image is created where the locations of green labels are plotted in white (255). The image represents the vegetation cover within the image.

4. **(Optional)** Perform a 'bitwise and' operation on the image using the grayscale image from step 3 as the mask. The resulting image shows the vegetation cover in the image within each of the segments of the PVC frame in the 3-channel RGB color space. The step, whose output is as shown in Figure 6(b), is merely to observe the vegetation detected within the image for user validation but is not required for analysis.

*C. Vegetation Cover to PVC Frame Segment Association*

Once the PVC frame and vegetation cover are identified, the next step in the process is to associate the vegetation cover with each segment of the PVC frame. The Hough Lines Transform is used to detect each of the segments of the PVC frame. The Hough Lines Transform is typically used to identify straight lines in images [5, 12]. Typically, straight lines are represented using the slope-intercept method as y = mx + c where m is the slope and c is the y-intercept. However,

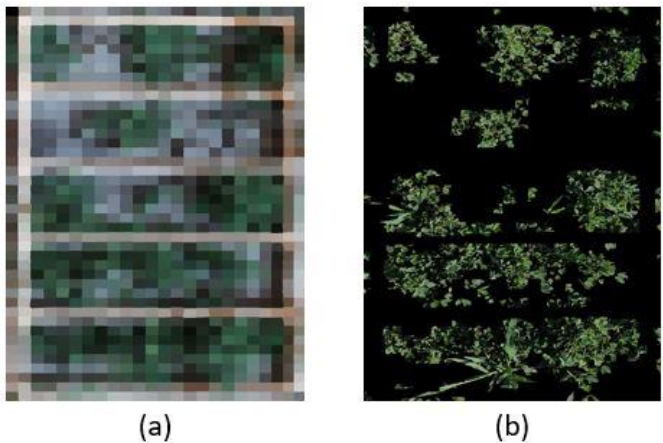

Figure 6: (a) Super-pixel Segmentation of Original Image

(b) Detected Vegetation Cover in RGB Color Space

Hough Transform relies on representing straight lines using the pair of polar co-ordinates, (ρ, θ). The first parameter, ρ, is the shortest distance from the origin to the line. The second parameter, θ, is the angle between the x-axis and distance line. The equation of the straight line in polar co-ordinate system is given by, ρ = x * cos(θ) + y * sin(θ) where (x, y) is a point on the line. The Hough Transform takes the edges from a binary image as input. The Hough Transform maps each of the pixels to multiple points in Hough (or parameter) space. An edge pixel is mapped to a sinusoid in 2D parameter space, (ρ, θ) representing all possible lines that could pass through the point. It is referred to as the voting stage. The sinusoids of collinear points in the Hough space cross each other indicating collinearity. There are two variants of Hough Transform, Standard and Probabilistic [17]. The primary difference between the two variants is the computational complexity. Consider an image that has M pixels as edge points and a Hough space divided in $N_ρ$ x $N_θ$ accumulators. In case of the Standard Hough Transform, each of the M edge pixels is used for computation which means that the computational complexity is $O(M. N_θ)$ for the voting stage and $O(N_ρ. N_θ)$ for the search stage. However, in the Probabilistic Hough Transform, only a subset m of M edge pixels is used for computation which means that the computational complexity is reduced to $O(m. N_θ)$ for the voting stage resulting in a faster execution of the algorithm.

The algorithmic steps to associate vegetation cover with PVC frame is described as follows.

1. Consider the image of the PVC frame obtained as the output in section *B*.

2. Detect straight lines on the image using the Probabilistic Hough Lines transform. The detected line segments indicate the segments of the PVC frame. A vote measure of 180 is used to start the search for straight lines. In case no lines are identified at the vote range, it is lowered by 10 until lines are detected.

3. Filter the line segments to identify the ones that run along the segments of the PVC frame. Any given segment in the PVC frame should have four lines with slope ~0.0 i.e. horizontal or nearly horizontal lines and two lines with infinite slope i.e. vertical or nearly vertical lines associated with it. In order to detect the lines, parallel lines with slopes less than 15 and slopes higher than 100 are retrieved. Lines with slopes under 15 are oriented at less than 10 degrees with the X-axis whereas lines with slopes above 100 approaching infinity are oriented at over 45 degrees with the X-axis. It is possible that more lines than expected are detected along the PVC frame segment. In order to contain only the expected six lines, any overlapping line segments or ones that are closer in position and orientation than a certain threshold are merged together into a single line segment. All line segments that are separated by less than 10 pixels in distance and 0.2 degree in orientation are grouped together and only a single line segment from amongst the group is considered for analysis.

4. Extend the line segments along the PVC frame segments by a factor of five (or higher) to ensure that the horizontal line segments intersect with the vertical line segments in case they do not.

5. Determine the segment of the PVC frame to which each of the line segments belong based on the position of the line segments in the image. In case of PVC frames that contain multiple segments, the association of line segments to the

PVC frame segments helps estimate the amount of vegetation cover within the PVC frame segments.

6. Consider the line segments associated with each segment of the PVC frame and determine the innermost lines i.e. two inner vertical and two inner horizontal lines. Find the four points of intersection of the innermost lines. The idea is to identify the inner rectangles (polygons) within the PVC frame.

7. Create a grayscale image and plot a polygon using the points of intersection as the four vertices of the polygon. Fill the polygon in white (255), as shown in Figure 7(b). In a PVC frame with n segments, n polygons are obtained where each of the polygons indicates a segment within the PVC frame.

8. Perform a 'bitwise and' operation on the output of section III(b), as shown in Figure 6(b), using each of the grayscale images as the mask. The pixels indicating vegetation cover are plotted in white (255), as shown in Figure 7(c).

### D. Amount of Vegetation Cover Estimation

The percentage of vegetation within each segment and the amount of vegetation cover in metric units are estimated by the algorithm. The ratio of the number of pixels that indicate vegetation cover and the number of pixels present in each segment of the PVC frame gives the percentage of vegetation cover within each PVC frame segment. In other words, the ratio of white pixels in images obtained from step 7 and step 8 gives the percentage of vegetation cover. The estimation of vegetation cover in metric units requires the knowledge of dimensions of the PVC frame. The length and width of the inner rectangles of the PVC frame shown in Figure 2(a) are 19.75" and 6.75" respectively. The number of pixels present within the rectangle are obtained from polygon masks plotted in step 7 of section III(C). Please note that each polygon might have slightly different pixel count since the image may be partly skewed. Since the number of vegetation cover pixels are known from step 8 of section III(C), the area of vegetation cover is given by the equation,

$$Vegetation\ Cover\ (sq.in) = (area\ of\ rectangle\ (sq.in)\ x\ number\ of\ vegetation\ cover\ pixels)/(number\ of\ polygon\ pixels).$$

### IV. RESULTS AND DISCUSSION

In the absence of ground truth measurements for the vegetation cover within each of the PVC frames, the proposed algorithm is evaluated on two applications SamplePoint and Canopeo. SamplePoint is a desktop application that facilitates foliar cover measurements from nadir imagery by superimposing a systematic or random array of up to 225 crosshairs targeting single image pixels and provides a platform for simple, manual classification of the pixels. SamplePoint is a product that has been developed in collaboration with the USDA and serves as the benchmark for vegetation cover estimation. Canopeo is a tool available as desktop and mobile applications to estimate fractional vegetation canopy cover. Both SamplePoint and Canopeo provide vegetation cover estimates as a percentage of the image size but not in metric units. It is perhaps due to the lack of a reference object in the image. The proposed algorithm is compared to the measurements provided by SamplePoint and Canopeo across 100 Kura plant images, similar to Figure 2(b). The Cosine Similarity [30] measure is used to compare similarity among the three measures. Cosine Similarity is the cosine of the angle between two vectors that are typically non-zero and within an inner product space. Mathematically, it is defined as the division between the dot product of vectors and product of the magnitude of each vector and is expressed as, $similarity = A.B/||A||\ ||B||$ where A and B are the two vectors compared for similarity. A plot of the estimates of vegetation cover across the 100 images using each of the three measures is shown in Figure 8. The cosine similarity measure between the proposed algorithm and SamplePoint is 0.995 whereas that between the proposed algorithm and Canopeo is 0.99. It is worth noting that the cosine similarity between SamplePoint and Canopeo is 0.99. The results indicate a high degree of cadence among the techniques and demonstrates the quality of results produced by the proposed algorithm. As observed from the median and mean values, the results produced by the proposed algorithm lie between that of the results produced by SamplePoint and Canopeo. The median of the proposed algorithm is 12.04 whereas that of SamplePoint and Canopeo are 13 and 9.14 respectively. The mean of the proposed algorithm is 12.40 whereas that of SamplePoint and Canopeo are 14.21 and 10.01 respectively. The key benefit to using the proposed technique is that the algorithm provides the measurement of the amount of vegetation cover in metric units along with percentage. It significantly improves reproducibility of the results across multiple individuals conducting the experiments since it eliminates the dependence of the results on the height from which the image is captured.

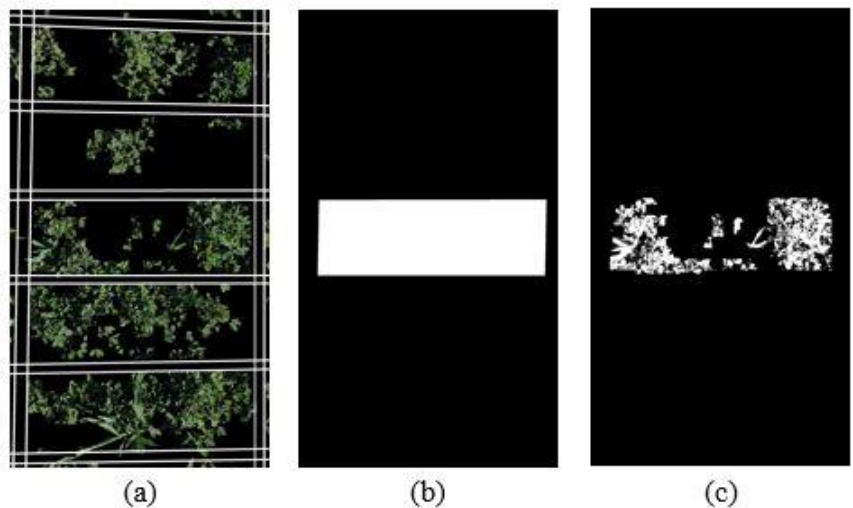

Figure 7: (a) Detected Hough Lines for the PVC Frame with Vegetation Cover (b) Polygon Mask for Segment 3 (c) Vegetation Cover within Segment 3

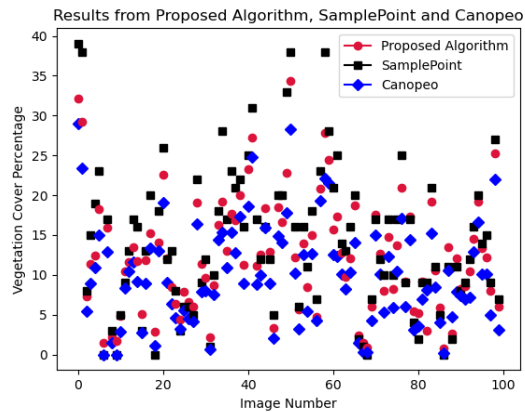

Figure 8: Vegetation Cover Estimates by the Proposed Algorithm, SamplePoint and Canopeo

## V. CONCLUSION AND FUTURE WORK

Proposed is an algorithm to estimate the FVC in an area from images to aid in Companion Planting. The proposed algorithm provides an estimate of the vegetation cover as a percentage of the image size and an absolute value in metric units. The algorithm relies on the presence of a PVC frame of known dimensionality in the image to be used as reference to measure the vegetation cover in metric units. The results of the proposed algorithm when compared to SamplePoint and Canopeo exhibit a high degree of similarity. Moving forward, the algorithm will be implemented as a mobile application that the users may download onto their devices. The code for the algorithm is also made open-source for enthusiasts to use and develop further. Any interest to collaborate is highly appreciated and invited.